\documentclass[11pt,a4paper]{article}
\usepackage{float}
\usepackage[hyperref]{acl2019}
\usepackage{times}
\usepackage{latexsym}
\usepackage{multirow}
\usepackage{graphicx}
\usepackage{amsmath}
\usepackage{comment}
\graphicspath{ ./} 

\usepackage{url}

\aclfinalcopy % Uncomment this line for the final submission
\def\aclpaperid{110} %  Enter the acl Paper ID here

\usepackage{flushend}

\title{From Bilingual to Multilingual Neural Machine Translation by Incremental Training} 

\author{Carlos Escolano,
        ~Marta R. Costa-juss\`a,
        ~Jos\'e A. R. Fonollosa,
        \\
  {\tt
  \{carlos.escolano,marta.ruiz,jose.fonollosa\}@upc.edu}\\
  TALP Research Center\\
  Universitat Polit\`ecnica de Catalunya, Barcelona\\
}

\date{}

\begin{document}
\maketitle

\begin{abstract}
Multilingual Neural Machine Translation approaches are based on the use of task-specific models and the addition of one more
language can only be done by retraining the whole system. In this work, we propose a new training schedule that allows the system to scale to more languages without modification of the previous components based on
joint training and language-independent encoder/decoder modules allowing for zero-shot translation. This work in progress shows close
results to the state-of-the-art in the WMT task.
\end{abstract}

\section{Introduction}\label{introduction}
%State of the art:
%\begin{itemize}
%\item Previous works share a single encoder and decoder to represent all languages in the same shared representation space \cite{johnson2017google}
%\item Another approach to the task is the use of specific encoders and decoders for each of the languages with the addition of a shared language independent layer\cite{lu2018neural}
%\item Related to the task are also important the recent work on unsupervised machine translation\cite{artetxe2017unsupervised, lample2017unsupervised} that combine the tasks of translation and autoencoding to compute a representation that is shared by the two languages.
%\end{itemize}
%
%
%Contributions:
%\begin{itemize}
%\item Multilingual model without parameter sharing
%\item A scalable multilingual model where new languages can be added without model retraining
%\item A new method of zero shot neural machine translation 
%\end{itemize}

In recent years, neural machine translation (NMT) has had an important improvement in performance. Among the different neural architectures, most approaches are based on an encoder-decoder structure and the use of attention-based mechanisms \citep{cho2014learning,bahdanau2014neural,vaswani2017attention}. The main objective is computing a representation of the source sentence that is weighted with attention-based mechanisms to compute the conditional probability of the tokens of the target sentence and the previously decoded target tokens. Same principles have been successfully applied to multilingual NMT, where the system is able to translate to and from several different languages. 

Two main approaches have been proposed for this task, language independent or shared encoder-decoders.  
Language independent architectures \cite{firat2016multi,firat2016zero,schwenk2017learning} in which each language has its own encoder and some additional mechanism is added to produce shared representations, as averaging of the context vectors or sharing the attention mechanism. These architectures have the flexibility that each language can be trained with its own vocabulary and all languages can be trained in parallel. Recent work \cite{lu2018neural} show how to perform many to many translations with independent encoders and decoders just by sharing additional language-specific layers that transform the language-specific representations into a shared one (without the need of a pivot language).

On the other hand, architectures that share parameters between all languages  \cite{johnson2017google} by using a single encoder and decoder are trained to be able to translate from and to any of the languages of the system. This approach presents the advantage that no further mechanisms are required to produced shared representation of the languages as they all share the same vocabulary and parameters, and by training all languages without distinction they allow low resources languages to take benefit of other languages in the system improving their performance. However, sharing vocabulary between all languages implies that the number of required tokens grows as more languages are included in the system, especially when languages employ different scripts in the system, such as Chinese or Russian. Recent work proposes a new approach to add new languages by adapting the vocabulary \cite{lakew2018multi} and relying on the shared tokens between languages to share model parameters. They show that the amount of shared tokens between languages has an impact in the model performance, which may be detrimental to languages with different scripts.

These approaches can be further explored into unsupervised machine translation where the system learns to translate between languages without parallel data just by enforcing the generation and representation of the tokens to be similar \cite{artetxe2017unsupervised,lample2017unsupervised}.

Also related to our method, recent work has explored transfer learning for NMT \cite{zoph-etal-2016-transfer, kim2019effective} to improve the performance of new translation directions by taking benefit of the information of a previous model. These approaches are particularly useful in low resources scenarios when a previous model trained with orders of magnitude more examples is available.

This paper proposes a proof of concept of a new multilingual NMT approach. The current approach is based on joint training without parameter  or vocabulary sharing by enforcing a compatible representation between the jointly trained languages and using multitask learning \cite{dong2015multi}. This approach is shown to offer a scalable strategy to new languages without retraining any of the previous languages in the system and enabling zero-shot translation. Also it sets up a flexible framework to future work on the usage of pretrained compatible modules for different tasks.

\section{Definitions}\label{definitions}
Before explaining our proposed model we introduce the annotation and background that will be assumed through the paper. Languages will be referred as capital letters $X,Y,Z$ while sentences will be referred in lower case $x,y,z$ given that $x\in X$, $y \in Y$ and $z \in Z$. 

We consider as an encoder ($e_x, e_y,e_z$) the layers of the network that given an input sentence produce a sentence representation ($h(x),h(y),h(z)$) in a space.  Analogously, a decoder ($d_x,d_y,d_z$) is the layers of the network that given the sentence representation of the source sentence is able to produce the tokens of the target sentence. Encoders and decoders will be always considered as independent modules that can be arranged and combined individually as no parameter is shared between them. Each language and module has its own weights independent from all the others present in the system.

\section{Joint Training}\label{joint}
%\begin{itemize}
% \item Independent modules
% \item No parameter sharing
% \item Joint training 
% \item Mesurable space distance
% \item Proposed distances: Correlation, "Comparative"
% \item Joint training autoencoder task and translation task to ensure that all modules are able to manage both languages.
% \item Failed distances: L1, adversarial distance. Space collapse
%\end{itemize}

In this section, we are going to describe the training schedule of our language independent decoder-encoder system. The motivation to choose this architecture is the flexibility to add new languages to the system without modification of shared components and the possibility to add new modalities (i.e. speech and image) in the future as the only requirement of the architecture is that encodings are projected in the same space. Sharing network parameters may seem a more efficient approach to the task, but it would not support modality specific modules. 

Given two languages, $X$ and $Y$, our objective is to train independent encoders and decoders for each language, $e_x, d_x$ and $e_y, d_y$ that produce compatible sentence representations $h(x), h(y)$. For instance, given a sentence $x$ in language $X$, we can obtain a representation $h(x)$ from the encoder $e_x$, than  can be used to either generate a sentence reconstruction using decoder $d_x$ or a translation using decoder $d_y$. %Additionally, we want that our architecture enables to add new languages without the need to retrain the current languages in the system.
With this objective in mind, we propose a training schedule that combines two tasks (auto-encoding and translation) and the two translation directions simultaneously by optimizing the following loss:

%\vspace{-1cm}
{\small
\begin{equation}
L = L_{XX} + L_{YY} + L_{XY} + L_{YX} + d
\end{equation}
%\vspace{-1cm}
}
where $L_{XX}$ and $L_{YY}$ correspond to the reconstruction losses of both language $X$ and $Y$ (defined as the cross-entropy of the generated tokens and the source sentence for each language); $L_{XY}$ and $L_{YX}$ correspond to the translation terms of the loss measuring token generation of each decoder given a sentence representation generated by the other language encoder (using the cross-entropy between the generated tokens and the translation reference); and $d$ corresponds to the distance metric between the representation computed by the encoders. This last term forces the representations to be similar without sharing parameters while providing a measure of similarity between the generated spaces.
We have tested different distance metrics such as L1, L2 or the discriminator addition (that tried to predict from which language the representation was generated). For all these alternatives, we experienced a space collapse in which all sentences tend to be located in the same spatial region. This closeness between the sentences of the same languages makes them non-informative for decoding. As a consequence, the decoder performs as a language model, producing an output only based on the information provided by the previously decoded tokens. Weighting the distance loss term in the loss did not improve the performance due to the fact that for the small values required to prevent the collapse the architecture did not learn a useful representation of both languages to work with both decoders.  To prevent this collapse, we propose a less restrictive measure based on correlation distance \cite{chandar2016correlational} computed as in equations \ref{eq:corr_dist} and \ref{eq:corr_dist2}. The rationale behind this loss is maximizing the correlation between the representations produced by each language while not enforcing the distance over the individual values of the representations.

\begin{comment}
{\small
\begin{gather} \label{eq:corr_dist}
 d = 1 - c(h(X),h(Y)) \nonumber \\ \\
 c(h(X),h(Y)) =  \nonumber \\ 
 \frac{\sum^{n}_{i=1}(h(x_{i} - \overline{h(X)}))(h(y_{i} - \overline{h(Y)}))}{\sqrt[]{\sum^{n}_{i}(h(x_{i})-\overline{h(X)})^{2}\sum^{n}_{i}(h(y_{i}) -\overline{h(Y)})^{2}}}  
\end{gather}
}%
\end{comment}

{\small
\begin{equation} \label{eq:corr_dist}
 d = 1 - c(h(X),h(Y))  
\end{equation} 
 }

{\small
\begin{gather} 
\label{eq:corr_dist2}
 c(h(X),h(Y)) =  \nonumber \\  
 \frac{\sum^{n}_{i=1}(h(x_{i} - \overline{h(X)}))(h(y_{i} - \overline{h(Y)}))}{\sqrt[]{\sum^{n}_{i}(h(x_{i})-\overline{h(X)})^{2}\sum^{n}_{i}(h(y_{i}) -\overline{h(Y)})^{2}}}  
\end{gather}
}%

%\begin{align*}
%c(h(X),h(Y)) =
%\end{align*}
%
%\begin{equation} \label{eq:corr_dist}
% \frac{\sum^{n}_{i=1}(h(x_{i} - \overline{h(X)}))(h(y_{i} - \overline{h(Y)}))}{\sqrt[]{\sum^{n}_{i}(h(x_{i})-\overline{h(X)})^{2}\sum^{n}_{i}(h(y_{i}) -\overline{h(Y)})^{2}}}
%\end{equation}

where $X$ and $Y$ correspond to the data sources we are trying to represent; $h(x_{i})$ and $h(y_{i})$ correspond to the intermediate representations learned by the network for a given observation; and $\overline{h(X)}$ and $\overline{h(Y)}$ are, for a given batch, the intermediate representation mean of $X$ and $Y$, respectively.

%\begin{eqnarray*}
%\left(1+x\right)ˆn & = & 1 + nx + \frac{n\left(n-1\right)}{2!}xˆ2 \\
%& & + \frac{n\left(n-1\right)\left(n-2\right)}{3!}xˆ3 \\
%& & + \frac{n\left(n-1\right)\left(n-2\right)\left(n-3\right)}{4!}xˆ4 \\
%& & + \ldots
%\end{eqnarray*}

\section{Incremental training}\label{scaling}
%\begin{itemize}
%    \item Add a new language without retraining or parallel corpus between all languages
%    \item Individual modules frozen
%    \item Allows using more data in low resources scenarios
%    \item zero shot machine translation
%\end{itemize}

%Additionally, we want that our architecture enables to add new languages without the need to retrain the current languages in the system.

Given the jointly trained model between languages $X$ and $Y$, the following step is to add new languages in order to use our architecture as a multilingual system. Since parameters are not shared between the independent encoders and decoders, our architecture enables us to add new languages without the need to retrain the current languages in the system. 
%As stated previously no parameter is shared between the different modules of our system and therefore each of them can be employed as a unit without affecting any of the other components.
Let's say we want to add language $Z$. To do so, we require to have parallel data between $Z$ and any language in the system. So, assuming that we have trained $X$ and $Y$, we need to have either $Z-X$ or $Z-Y$ parallel data. For illustration, let's choose to have $Z-X$ parallel data. Then, we can set up a new bilingual system with language $Z$ as source and language $X$ as target. To ensure that the representation produced by this new pair is compatible with the previously jointly trained system, we use the previous $X$ decoder ($d_x$) as the decoder of the new $ZX$ system and we freeze it. During training, we optimize the cross-entropy between the generated tokens and the language $X$ reference data but only updating the layers belonging to the language $Z$ encoder ($e_z$). Doing this, we train $e_z$ not only to produce good quality translations but also to produce similar representations to the already trained languages. No additional distance is added during this step. The language $Z$ sentence representation $h(z)$ is only enforced by the loss of the translation to work with the already trained module as it would be trained in a bilingual NMT system.

%However, if data in both training steps are extracted from different domains, the system may produce longer sequences by encoding the data as shorter pieces compared to the previously trained system.

\begin{figure}[h]
\begin{center}
\includegraphics[scale=0.40]{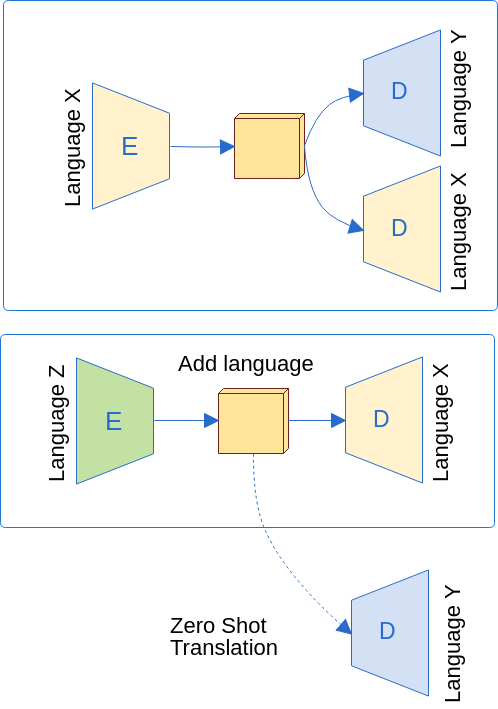}
\caption{Language addition and zero shoot training scheme}
\label{img:scalability}
\end{center}
\vspace{-0.75cm}
\end{figure}

Our training schedule enforces the generation of a compatible representation, which means that the newly trained encoder $e_z$ can be used as input of the decoder $d_y$ from the jointly trained system to produce zero-shot $Z$ to $Y$ translations. See Figure \ref{img:scalability} for illustration.

The fact that the system enables zero-shot translation shows that the representations produced by our training schedule contain useful information and that this can be preserved and shared to new languages just by enforcing the new modules to train with the previous one, without any modification of the architecture. Another important aspect is that no pivot language is required to perform the translation, once the added modules are trained the zero-shot translation is performed without generating the language used for training as the sentence representations in the shared space are compatible with all the modules in the system.

A current limitation is the need to use the same vocabulary for the shared language ($X$) in both training steps. The use of subwords \cite{sennrich2015neural} mitigates the impact of this constraint.

%Given an $X$-$Y$ system jointly trained to translate between languages $X$ and $Y$ in both directions and a parallel corpus of languages $X$ and $Z$. In order to add to the system the translation direction $Z$-$X$ we can train a new language $Z$ encoder to produce a compatible representation with the ones produced by the jointly trained system.
%
%This encoder is trained as a Transformer model with the same hyperparameters of the jointly trained system, having the decoder the same weights of the language $X$ decoder. By freezing the decoder weights we can train the decoder as a bilingual Transformer model by optimizing the cross-entropy of the generated tokens.

%\section{Training curriculum}\label{curriculum}
%\input{sections/curriculum.tex}

\section{Data and Implementation}\label{data}
%\begin{itemize}
%\item BPE encoding for all languages. 32000 operations and not shared code files
%\item WMT data for Spanish, English, French and German. Sources? 18 million sentences.
%\item Dev set: newstest2012
%\item Test set: newstest2013
%\item Dev and Test sets parallels between all the languages (to test the zeroshot translation)
%\end{itemize}

% \usepackage{multirow}
Experiments are conducted using data extracted from the UN \cite{ziemski2016united} and EPPS datasets \cite{koehn2005europarl} that provide 15 million parallel sentences between English and Spanish, German and French. \textit{newstest2012} and \textit{newstest2013} were used as validation and test sets, respectively. These sets provide parallel data between the four languages that allow for zero-shot evaluation.
 Preprocessing consisted of a pipeline of punctuation normalization, tokenization, corpus filtering of longer sentences than 80 words and true-casing. These steps were performed using the scripts available from Moses \cite{koehn2007moses}. Preprocessed data is later tokenized into BPE subwords \citep{sennrich2015neural} with a vocabulary size of 32000 tokens. We ensure that the vocabularies are independent and reusable when new languages were added by creating vocabularies monolingually, i.e. without having access to other languages during the code generation.

{
\begin{table}[]
\small
\begin{tabular}{|l|llll|}
\hline
\textbf{System} & \textbf{ES-EN} & \textbf{EN-ES} & \textbf{FR-EN} & \textbf{DE-EN}  \\ \hline
Baseline        & 32.60          & 32.90          & 31.81          & 28.96            \\
Joint & 29.70          & 30.74          & -              & -                                  \\
Added lang  & -              & -              & 30.93          & 27.63                  \\ \hline
\end{tabular}
\caption{Experiment results measured in BLEU score. All blank positions are not tested or not viable combinations with our data.}
\label{table:results}
\end{table}}

\begin{table}[]
\begin{center} 
\small
\begin{tabular}{|l|ll|}
\hline
\textbf{System} & \textbf{FR-ES} & \textbf{DE-ES} \\ \hline
%& \textbf{DE\_l-ES} \\ \hline
Pivot           & 29.09          & 21.74     \\   
%& 15.71             \\
Zero-shot       & 19.10          & 10.92        \\ \hline  
%& 6.96              \\ \hline
\end{tabular}
\caption{Zero-shot results measured in BLEU score}
\label{results-zeroshot}
\end{center}
\vspace{-0.5cm}
\end{table}

\begin{figure*}[t]
\begin{center}
\includegraphics[scale=0.38]{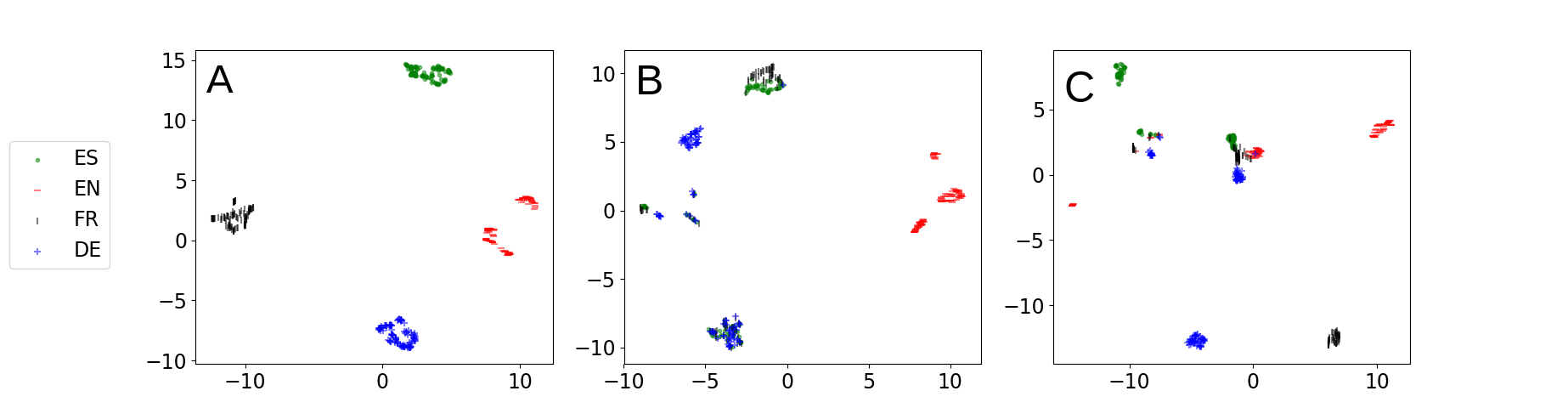}
\caption{Plot A shows the source sentence representation of each of the encoder modules(ES,EN,DE,FR). Plots B and C show the representation of the target sentence generated by English(B) and Spanish(C) decoders given the sentence encodings of parallel sentences generated for all four language encoder modules.}
\label{img:vis-encoder} 
\end{center}
\vspace{-0.5cm}
\end{figure*}

\begin{comment}
\begin{figure*}[t]
\begin{center}
\includegraphics[scale=0.26]{images/layers-es.png}}
\caption{Sentence representation at the first, third and last layer of the Spanish decoder for 200 examples of parallel sentences in Spanish and English. Plots generaters as explained in \cite{lacroux2019}}
\label{img:layers-decoder}
\end{center}
\vspace{-0.5cm}
\end{figure*}
\end{comment}

\section{Experiments}\label{experiments}
%All models trained with the following parameters:
%\begin{itemize}
%\item Transformer architecture (based on Fairseq implementation)
%\item 6  blocks encoder and decoder
%\item Hidden size and embedding size 512
%\item 4 heads attention 
%\item Update weights every 16 batches
%\end{itemize}
%
%Joint training on 2 GPUS Nvidia Titan X. Added languages on 1 gpu.

%In this section we are going to discuss the results of our proposed architecture into different tasks, train a system to translate between similar languages highly resourced and train a system for limited resources with not related languages and different scripts (Latin and Cyrillic).

%\subsection{General Case. Similar Resources Languages}

Our first experiment consists in comparing the performance of the jointly trained system to the standard Transformer. As explained in previous sections, this joint model is trained to perform two different tasks, auto-encoding and translation in both directions. In our experiments, these directions are Spanish-English and English-Spanish. In auto-encoding, both languages provide good results at 98.21 and 97.44 BLEU points for English and Spanish, respectively. In translation, we observe a decrease in performance.  Table \ref{table:results} shows that for both directions the new training performs more than 2 BLEU points below the baseline system. %Even though the objectives of both systems are not exactly the same. 
This difference suggests that even though the encoders and decoders of the system are compatible they still present some differences in the internal representation.

Note that the languages chosen for the joint training seem relevant to the final system performance because they are used to define the representations of additional languages. Further experimentation is required to understand such impact.

%To have a better understanding of how the languages involved in the system affect its performance 
Our second experiment consists of incrementally adding different languages to the system, in this case, German and French. Note that, since we freeze the weights while adding the new language, the order in which we add new languages does not have any impact on performance. Table \ref{table:results} shows that French-English performs 0.9 BLEU points below the baseline and German-English performs 1.33 points below the baseline. French-English is closer to the baseline performance and this may be due to its similarity to Spanish, one of the languages of the initial system languages.  %The lower performance in German may be due to the fact that German can combine several words into a single word (agglutinative) and it is not a Latin language like the others present in the system (Spanish and French).

%Additionally to new languages we tried to measure the impact of the amount of data in the relative performance of the system when a new language was added. Our experiment with a subset of the German data of half a million sentences shows a similar BLEU difference compared to the German encoder trained with all the data available. Results show a difference of 1.66 points, suggesting that the encoder is not taking advantage of the fact that the decoder was trained with more data during the joint training.

The added languages have better performance than the jointly trained languages (Spanish-English from the previous section).  This may be to the fact that the auto-encoding task may have a negative impact on the translation task.

%\subsection{Zero-shot translation}

Finally, another relevant aspect of the proposed architecture is enabling zero-shot translation. To evaluate it, we compare the performance of each of the added languages compared to a pivot system based on cascade. Such a system consists of translating from French (German) to English and from English to Spanish with the standard Transformer. Results show that the zero shot translation provides a consistent decrease in performance for both cases of zero-shot translation.

\section{Visualization}\label{visualization}
Our training schedule is based on training modules to produce compatible representations, in this section we want to analyze this similarity at the last attention block of encoders, where we are forcing the similarity. In order to graphically show the presentation a UMAP \cite{mcinnes2018umap-software} model was trained to combine the representations of all languages. Figures \ref{img:vis-encoder} (A), (B) and (C) show 130 sentences extracted from the test set. These sentences have been selected to have a similar length to minimize the amount of padding required.

\begin{comment}
\begin{figure}
\begin{center}
\includegraphics[scale=0.40]{images/plot_encoders.png}}
\caption{Encoder representations}
\label{img:vis-encoder}
\end{center}
\end{figure}
\end{comment}

Figure \ref{img:vis-encoder} (A) shows the representations of all languages created by their encoders. Languages tend to be represented in clusters with no complete overlapping between languages as we would have liked. This mismatch in the intermediate representation is similar to what \cite{lu2018neural} reported in their multilingual approach, where authors argue that the language dependent features of the sentences have a big impact in their representations. 
%The separated clusters shows that languages are not yet represented in the same space. %However, the added languages using the English decoder present a more similar shape between them and with English than the Spanish ones.
However, since our encoder/decoders are compatible and produce competitive translations, we decided to explore the representations generated at the last attention block of the English decoder, and they are shown in Figure \ref{img:vis-encoder} (B). We can observe much more similarity between English, French, and German, (except for a small German cluster) and separated clusters for Spanish. The reason behind these different behaviors may be due to the fact that French and German have directly been trained with the frozen English decoder and being adjusted to produce representations for this decoder. 
%Even between French and German we can observe how while French is always close to  English there is a German cluster distant from the rest. 
Finally, Figure \ref{img:vis-encoder} (C) shows the representations of the Spanish decoder. Some sentences have the same representation for all languages, whereas others no. 
Looking at the specific sentences that are plotted, we found that close representations do not correlate with better translations or better BLEU (as shown in examples from the Appendix \ref{a:data}). More research is required to analyze which layer in the decoder is responsible for approaching languages in a common space. This information could be used in the future to train encoders of new languages by wisely sharing parameters with the decoder as in previous works \cite{he:2018}. %, while the difference for French-English is 0.9 points for Spanish it increases to 2.9 points which is also the most dissimilar language. 

\section{Conclusions}\label{conclusions}
%\begin{itemize}
%\item Scalable model without retraining
%\item Comparable results to the baseline model in the supervised tasks
%\item Zeroshot learning. Worst performance than pivot system
%\end{itemize}

This work proposes a proof of concept of a bilingual system NMT which can be extended to a multilingual NMT system by incremental training. We have analyzed how the model performs for different languages. Even though the model does not outperform current bilingual systems, we show
first steps towards achieving competitive translations with a flexible architecture that enables scaling to new languages (achieving multilingual and
zero-shot translation) without retraining languages in the system.

%Even though the model does not outperform current bilingual systems, we show first steps towards achieving competitive translations with a flexible architecture that enables scaling to new languages (achieving multilingual and zero-shot translation) without retraining languages in the system. 

\section*{Acknowledgments}

This work is supported in part by a Google Faculty Research Award.
This work is also supported in part by the
Spanish Ministerio de Economía y Competitividad,
the European Regional Development Fund
and the Agencia Estatal de Investigación,
through the postdoctoral senior grant Ram\'on y Cajal, contract TEC2015-69266-P
(MINECO/FEDER,EU) and contract PCIN-2017-079 (AEI/MINECO).

\bibliography{acl2019}
\bibliographystyle{acl_natbib}

\appendix

\section{Examples}
\label{a:data}

This appendix shows some examples of sentences visualized in Figure 2. Table 1 reports outputs produced by the Spanish decoder given encoding representations produced by the Spanish, English, French and German encoder.  The first two sentences have similar representations between the languages
in Figure 2 (C) (in the Spanish decoder visualization).  While the first one keeps the meaning of the sentence,
the second one produces meaningless translations. The third sentence produces disjoint representations but the
meaning is preserved in the translations. Therefore, since close representations may imply different translation performance, further research is required to understand the correlation between representations and translation quality.

\begin{table*}[h]
\begin{center}
\small
\begin{tabular}{|p{1.5cm}|p{12cm}|}
\hline
\textbf{System} & \textbf{Sentence}  \\ \hline
Reference       &    ponemos todo nuestro empe\~no en participar en este proyecto .                 \\
ES              &    ponemos todo nuestro empe\~no en participar en este proyecto .                 \\
EN              &    participamos con esfuerzo en estos proyctos .         \\
FR              &    nos esfuerzos por lograr que los participantes intensivamente en estos proyectos.       \\
DE              &    nuestro objetivo es incorporar estas personas de manera intensiva en nuestro proyecto.           \\ \hline
Reference       &    ¡Caja Libre!               \\
ES              &    ¡Caja Libre|                 \\
EN              &    Free chash points!     \\
FR              &    librecorrespondinte.        \\
DE              &    cisiguinente           \\ \hline
Reference       &    ¿C\'omo aplica esta definici\'on en su vida cotidiana y en las redes sociales?    \\
ES              &    ¿C\'omo aplica esta definici\'on en su vida cotidiana y en las redes sociales?               \\
EN              &    ¿C\'omo se aplica esta definici\'on a su vida diaria?    \\
FR              &    ¿C\'omo aplicar esta definici\'on en la vida diaria y sobre los red sociales?       \\
DE              &    ¿Qu\'e es aplicar este definici\'on a su dadadato y las redes sociales?           \\ \hline
\end{tabular}
\caption{Outputs produced by the Spanish decoder given encoding representations produced by the Spanish, English, French and German encoder.}
\end{center}
\end{table*}

\begin{table*}[h]
\begin{center}
\small
\begin{tabular}{|p{1.5cm}|p{12cm}|}
\hline
\textbf{System} & \textbf{Sentence}  \\ \hline
Reference       &    it was a terrific season.                \\
ES              &    we had a strong season .                \\
EN              &    it was a terrific season.\\
FR              &    we made a very big season .      \\
DE              &    we have finished the season with a very strong performance.           \\ \hline
Reference       &    in London and Madrid it is completely natural for people with serious handicaps to be independently out in public, and they can use the toilets, go to the museum, or wherever ...               \\
ES              &    in London and Madrid , it is very normal for people with severe disability to be left to the public and be able to serve , to the museum , where ...
             \\
EN              &     in London and Madrid it is completely natural for people with serious handicaps to be independently out in public, and they can use the toilets, go to the museum, or wherever ...      \\
FR              &   in London and Madrid, it is quite common for people with a heavy disability to travel on their own in public spaces; they can go to the toilets, to the museum, anywhere ...       \\
DE              &     in London and Madrid, it is absolutely common for people with severe disabilities to be able to move freely in public spaces, go to the museum, use lets, etc.     \\ \hline
Reference       &    from the Czech viewpoint, it seems they tend to put me to the left.    \\
ES              &    from a Czech point of view, I have the impression that people see me more than on the left.               \\
EN              &    from the Czech viewpoint, it seems they tend to put me to the left.   \\
FR              &    from a Czech point of view , I have the impression that people are putting me on the left .     \\
DE              &    from a Czech point of view, it seems to me that people see me rather on the left.         \\ \hline
\end{tabular}
\caption{Outputs produced by the English decoder given encoding representations produced by the Spanish, English, French and German encoder.}
\end{center}
\end{table*}

Table 2 shows outputs produced by the English decoder given encoding representations produced by the Spanish, English, French and German encoder. All examples appear to be close in Figure 2 (B) between German, French and English. We see that the German and French outputs preserve the general meaning of the sentence. Also and differently from previous Table 1, the outputs do not repeat tokens or produce unintelligible translations. For these cases, there are not sentences from French that appear far away in the visualization, so again, we need further exploration to understand the information of these intermediate representations. Just recently, we have released our visualization tool \cite{demopaper}.

\end{document}